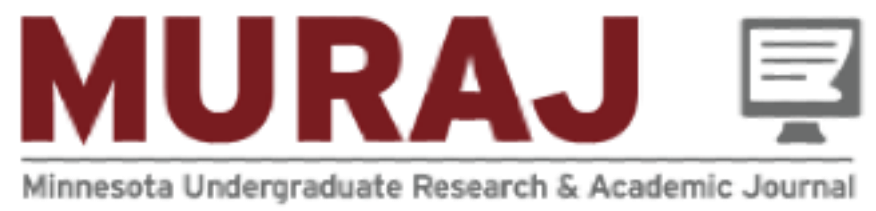

# An Exploratory Study of Single Channel Surface Electromyography for Hand Gesture Classification

**Daanish Hindustani**[1]

## Abstract

Accurate hand gesture recognition using surface electromyography (sEMG) typically relies on multichannel sensor arrays and computationally intensive models. This limits practical deployment in low power and embedded systems. This study investigates the feasibility of classifying ten hand gestures using a single sEMG channel combined with lightweight machine learning architectures. Raw sEMG signals were transformed into a comprehensive feature-based representation, including time domain, frequency domain, higher order crossing, and relative intensity features. Feature redundancy was reduced using Pearson correlation filtering and the removal of highly correlated features, while dimensionality reduction techniques (LDA and PCA) were applied selectively. Three classifiers—feed forward neural network (NN), k-nearest neighbors (KNN), and support vector machine (SVM)—were systematically evaluated across four experiments. Results demonstrate that combining time and frequency features with Pearson filtering and a compact NN can achieve up to 90% accuracy, even with limited temporal and spatial information. These findings highlight the potential for single channel sEMG systems in cost-effective, low-power, gesture-recognition applications.

## 1 Introduction

Hand gestures are generated by coordinated contractions of forearm and hand muscles, which actuate finger bones (phalanges) through tendons. These muscle activations are driven by neural signals originating in the central nervous system and can be noninvasively captured using surface electromyography (sEMG) [8]. sEMG signals provide a quantitative representation of muscle activity and have been widely adopted for hand gesture recognition in applications such as prosthetic control, rehabilitation, and human–computer interaction (HCI) [1], [3], [8].

Recent advances in machine learning have enabled high- accuracy gesture classification using dense sEMG sensor ar- rays with up to 128–256 channels distributed around the forearm [2], [10]. While effective, such systems are often impractical for large-scale or consumer deployment due to increased hardware complexity, power consumption, and cost [2]. Furthermore, models trained on high-dimensional, multi- channel data require substantial computational resources, making

[1] Department of Computer Science, University of Minnesota

real-time inference on embedded platforms challenging [5].

This research investigates whether accurate hand gesture classification can be achieved using a single sEMG channel and a compact classification model suitable for deployment on low-power devices such as a Raspberry Pi. Motivated by recent work on single-channel and embedded sEMG systems [9], [11], the proposed pipeline leverages feature engineering from the time and frequency domains, correlation-based filtering, and dimensionality reduction techniques to reduce the number of channels without sacrificing performance.

We conducted four structured experiments, each comprising of three sub-experiments to evaluate the effects of different preprocessing pipelines and classifiers. The results demonstrate that specific features, combined with a small neural network, can achieve competitive accuracy even with reduced temporal and spatial information. [1], [10].

## 2 Related Work

Single channel sEMG gesture recognition research builds on foundational work by Khushaba et al. [8], which established a benchmark dataset for 10 finger movements using flexor channel signals from 8 subjects. Their study recorded sEMG at 4000 Hz from Delsys DE-2.1 sensors on the volar forearm, extracting time-domain (MAV, RMS, WL) and TD-PSD fre- quency features from 300ms windows. Combining mutual information and genetic algorithms, they achieved 93% accuracy with a quadratic discriminant on just two channels. This demonstrates flexor channel dominance for discriminative gesture patterns [8]. Marba´n Salgado et al. [11] advanced single-channel deploy- ment by implementing a feed forward ANN directly on an ESP32 microcontroller for real-time prosthetic control. Their pipeline processed flexor sEMG at 1000 Hz, extracting 7 time- domain features (MAV, RMS, WL, ZC, SSC, WAMP, MAX) from 250ms windows with 50ms overlap. After PCA reduction to 4 components, a compact ANN (input-16-8-10) achieved 92.3% accuracy across 8 gestures using only 240KB RAM and 1.2ms inference time. Notably, their Pearson correlation threshold of 0.85 retained 68% of features while eliminating redundancy, mirroring the filtering strategies evaluated here but optimized for 32-bit MCU constraints rather than Raspberry Pi capabilities [11].

Additional single-channel studies validate feature engineering approaches. Li et al. [9] reported 97.5% accuracy for binary open/close classification using KNN on 12 time-domain features from a single forearm channel, though limited to two gestures. More recent work by Al-Timemy et al. [2] reviewed single channel sEMG classification and achieved 85-92% accuracy for 6-10 gestures. This success is due to combining time/frequency features with LDA/PCA. However, the dimensionality reduction led to a 10-15% drop in spatial information loss.

This study extends these works by leveraging Pearson filtering (0.1 threshold), correlation filtering (95%), LDA (9 components), and PCA (5 components) across NN/KNN/SVM on Khushaba's full 10-gesture dataset. Unlike Marba´n's ESP32 focus, the evaluation targets Raspberry Pi deployment with larger architectures (up to 256 neurons), achieving 90% ac- curacy that rivals embedded systems while enabling more complex gesture discrimination [1], [11].

# 3 Methods

## 3.1 Data Collection

The dataset used in this study was obtained from the work of Khushaba et al. [8], which investigated surface elec- tromyogram (sEMG) signals for improved control of prosthetic fingers. Eight healthy subjects (six male and two female), aged between 20 and 35 years, participated in the data collection. All subjects were normally limbed and reported no neurological or muscular disorders. Prior to participation, informed consent was obtained from all subjects in accordance with the experimental protocol described in [8].

During data acquisition, subjects were seated on an armchair with the forearm supported and fixed in a single position to minimize the effect of limb position variability on the recorded EMG signals. EMG data were collected using two surface EMG channels (Delsys DE 2.x series sensors) and processed using a Bagnoli Desktop EMG System (Delsys Inc.). A two slot adhesive skin interface was used to securely attach each sensor to the skin, and a conductive adhesive reference electrode was placed on the wrist of each subject. The electrode placement locations on the volar (flexor - Channel 1) and dorsal (extensor - Channel 2) sides of the right forearm are illustrated in Figure 1.

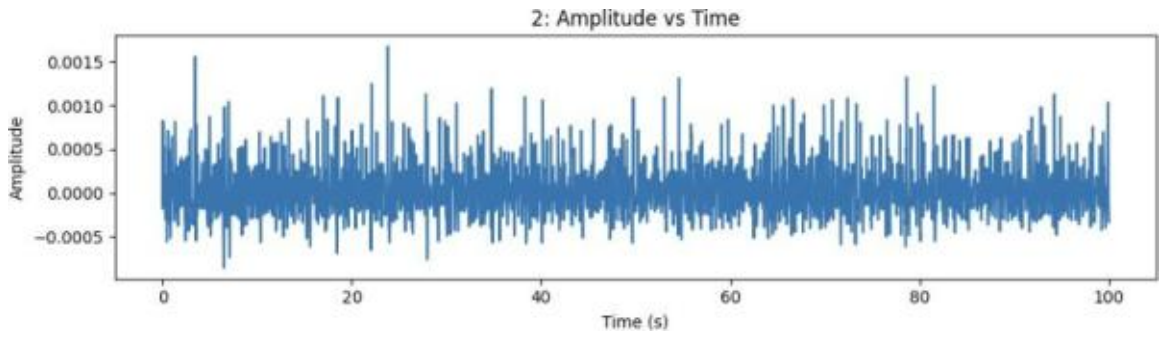

*Figure 1. Amplitude versus time representation of the single channel sEMG signal recorded from the middle finger.*

The EMG signals were amplified using a Delsys Bagnoli 8 amplifier with a total gain of 1000. Signals were digitized using a 12-bit analog-to-digital converter (National Instruments BNC 2090) at a sampling rate of 4000 Hz and acquired using Delsys EMGWorks Acquisition software. Following acquisition, the signals were bandpass filtered between 20–450 Hz, with an additional notch filter applied to suppress 50 Hz power line interference, consistent with the original experimental setup in [8].

Ten finger movement classes were recorded. These included individual finger flexions: Thumb (T), Index (I), Middle (M), Ring (R), and Little (L), as well as combined finger pinching gestures: Thumb Index (T–I), Thumb Middle (T–M), Thumb Ring (T–R), Thumb Little (T–L), and Hand Close (HC). Example electrode placements and gesture illustrations are shown in Figures 2 and 3, respectively.

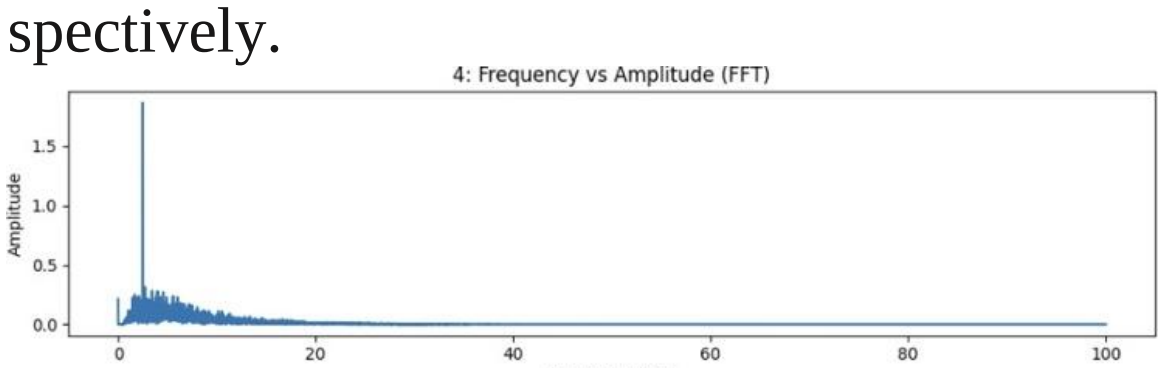

*Figure 2. Time domain waveform of the sEMG signal recorded from the middle finger.*

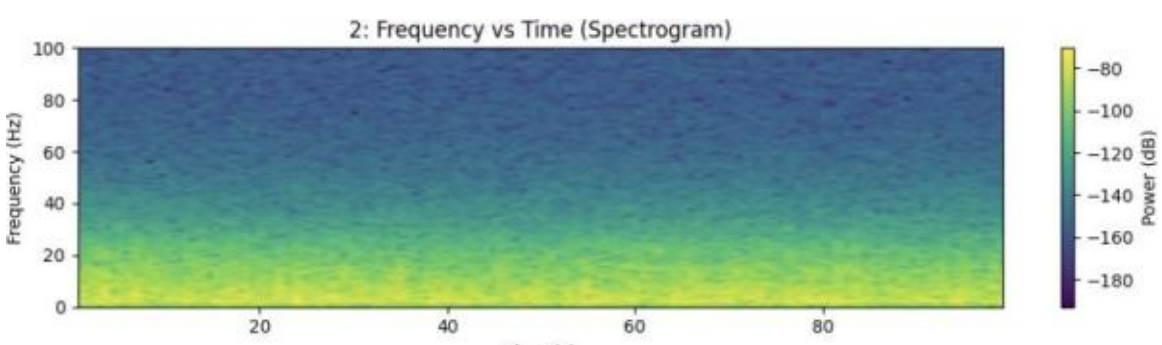

*Figure 3. Time frequency spectrogram of the sEMG signal recorded from the middle finger.*

The experimental protocol originally conducted in [8] prompted subjects, via an auditory cue, to transition from a resting state to a target gesture and maintain the contraction for 5 seconds. Each gesture was repeated six times, with a resting period of 3 to 5 seconds between trials. Although two EMG channels are available in the dataset, only Channel 1, representing the sEMG signals acquired from the volar (flexor) muscles, was used to evaluate the feasibility of single-channel hand gesture classification. This is because the flexor muscles are known to produce stronger and more discriminative sEMG signals for hand gesture recognition [8].

### 3.2 Feature Extraction

Raw sEMG signals as direct inputs to the classification model are both computationally expensive and storage intensive. Therefore, this study employs a feature-based representation. Raw sEMG signals are inherently high dimensional, noisy, and highly variable across subjects and recording sessions, making them difficult for models to interpret directly without extensive preprocessing or deep architectures [3], [7], [8]. Additionally, the high sampling rate of 4000 Hz generates a large volume of data, further increasing memory requirements and training time.

To address these challenges, the raw signals were segmented into overlapping windows of 300 ms with a 50 ms overlap. Feature extraction was then performed on each valid window. A total of 32 features were extracted per window, yielding approximately 57,000 samples. These features span multiple domains and were selected based on prior literature for their effectiveness in capturing discriminative characteristics of sEMG signals [1], [7], [8]:

Time-domain features: Integrated EMG (IEMG), Mean Absolute Value (MAV), Root Mean Square (RMS), Variance (VAR), Waveform Length (WL), Zero Crossings (ZC), Simple Square Integral (SSI), Log Detector (LOG), Willison Amplitude (WAMP), Maximum Fractal Length (MFL), Slope Sign Changes (SSC), Maximum Value (MAX) [3].

Autoregressive features: AR coefficients (AR1-AR4) [8]. Higher order crossing features: Mean of Amplitude and Derivative Order Crossings (MOAC aa, MOAC ad, MOAC da, MOAC dd), Amplitude of Power Order Cross- ings (APOC aa, APOC ad, APOC da, APOC dd), Stan- dard Deviation of Order Crossings (STDOC aa, STDOC ad, STDOC da, STDOC dd) [7]. Relative intensity features: RI a mean, RI d mean, RI aad mean [1]. Frequency-domain feature: Yule–Burg Spectrum (YBS) [7].

This carefully chosen feature set balances expressiveness and computational efficiency, enabling effective classification while significantly reducing input dimensionality compared to raw signals. By transforming the raw sEMG into a compact, informative representation, the model can focus on discriminative patterns relevant to hand gestures rather than being overwhelmed by raw signal noise and redundancy [3], [8].

### 3.3 Data Pre-processing

Two correlation-based preprocessing techniques were applied to reduce feature redundancy and improve model per- formance, as commonly used in sEMG and machine learning studies [6], [7]:

Correlation filtering: Features with pairwise correlation exceeding 95% were removed. This threshold was chosen due to the high redundancy observed among extracted features (Figure 4) [6].

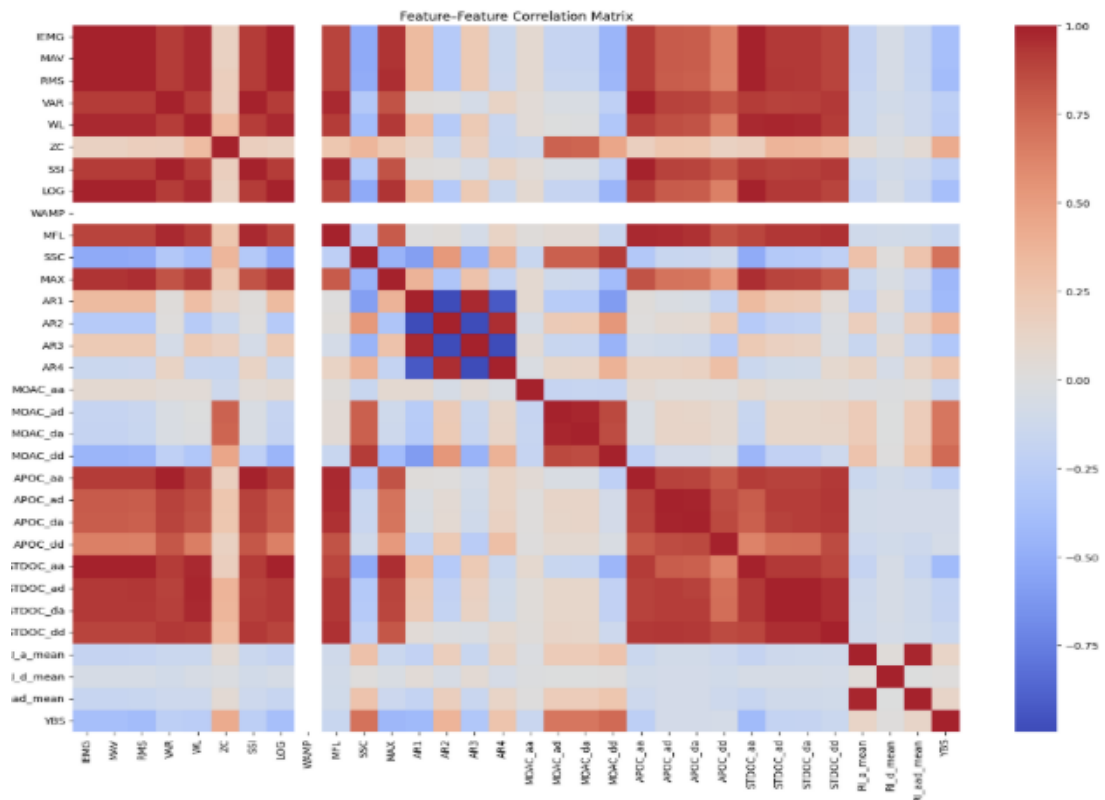

*Figure 4. Inter feature correlation matrix highlighting redundancy among extracted features prior to dimensionality reduction.*

Pearson filtering: Features with an absolute Pearson correlation below 0.1 with respect to the class labels were removed, eliminating weakly informative features (Figure 5) [7].

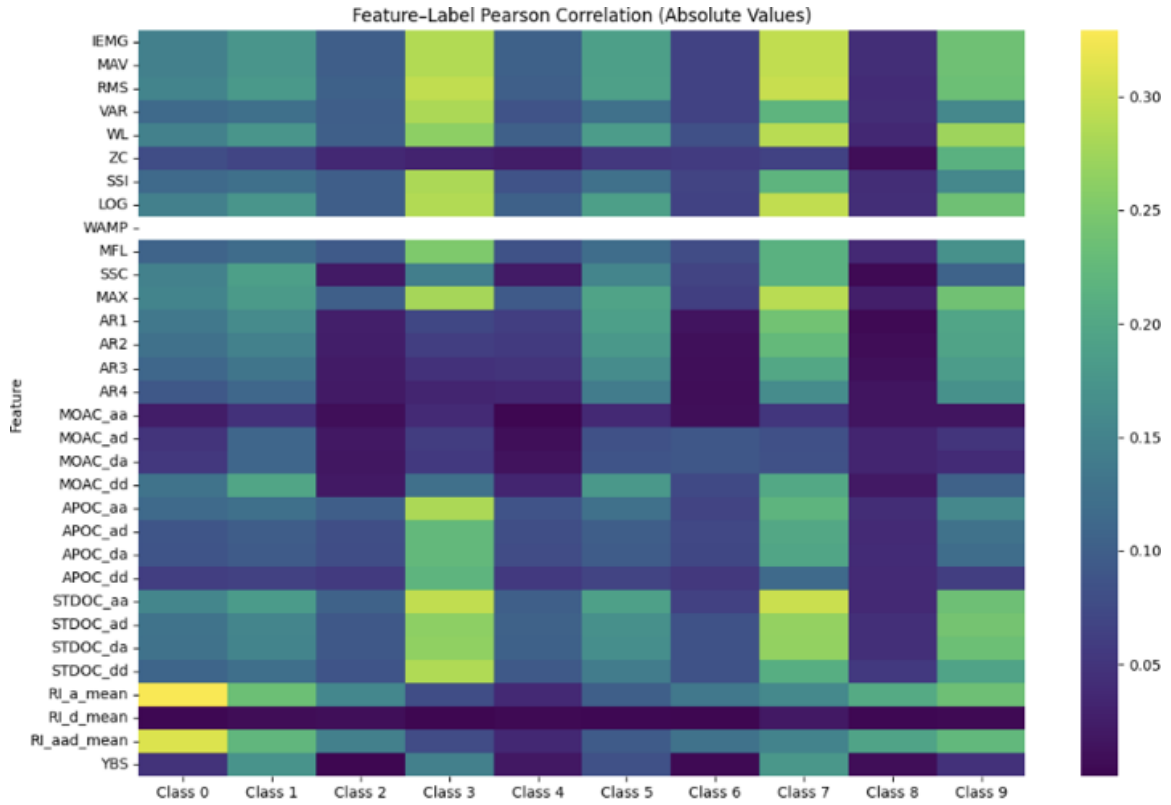

*Figure 5. Pearson correlation matrix for the complete set of 32 extracted time and frequency domain features.*

These methods were used individually or in combination across experiments and served as initial dimensionality reduction steps prior to classification [6], [7].

# 4 Results

To systematically evaluate the performance of singlechannel sEMG classification, we designed a series of exper- iments. Each experiment represents a unique combination of feature preprocessing, dimensionality reduction, and classifier type. The preprocessing steps, including Pearson correlation filtering, general correlation filtering, Linear Discriminant Analysis (LDA), and Principal Component Analysis (PCA), were selectively applied and varied across experiments. This was to isolate their individual and combined effects on classification performance [6], [7].

For each experimental condition, one of three classifiers was used as a sub-experiment: a feed forward neural network (NN), k-nearest neighbors (KNN), or a support vector machine (SVM). This structure allows us to systematically compare the impact of preprocessing and dimensionality reduction techniques across different model types. This will lead to the optimal configuration for real world deployment [4], [12].

All experiments used the same dataset described in Section 3A. Then the feature extraction followed the pipeline outlined in Section 3B, resulting in a comprehensive set of 32 features per window. By varying preprocessing and dimensionality reduction for each sub experiment, we ensure that the effects of each technique are independently measurable and interpretable [8].

## 4.1 Experiment 1 and Results

In Experiment 1, the data was preprocessed using both the Pearson correlation filter and a general correlation filter, reducing the dimensionality to 8 features. The dataset was then split into 80% training and 20% testing sets. The training set was normalized, and the same normalization parameters were applied to the test set to avoid data leakage[4].

1) Neural Network (NN): The first sub-experiment employed a feed-forward neural network trained on the preprocessed dataset, which was converted into tensor form. The network architecture consisted of three hidden layers with 32, 64, and 32 neurons, respectively, in addition to the input and output layers. This architecture was chosen empirically; larger networks led to overfitting and reduced generalization [11]. The model used ReLU activation functions, a batch size of 32, the Adam optimizer with a learning rate of 0.001, and was trained for 100 epochs [4].

This configuration achieved an average accuracy and F1 score of 81%. Training loss plateaued around 20% while the testing accuracy was around 80%. This suggests that the selected feature set and network capacity were insufficient to fully distinguish between certain hand gesture classes. The confusion matrix in Figure 7 shows that gestures 3, 4, and 7 were frequently misclassified.

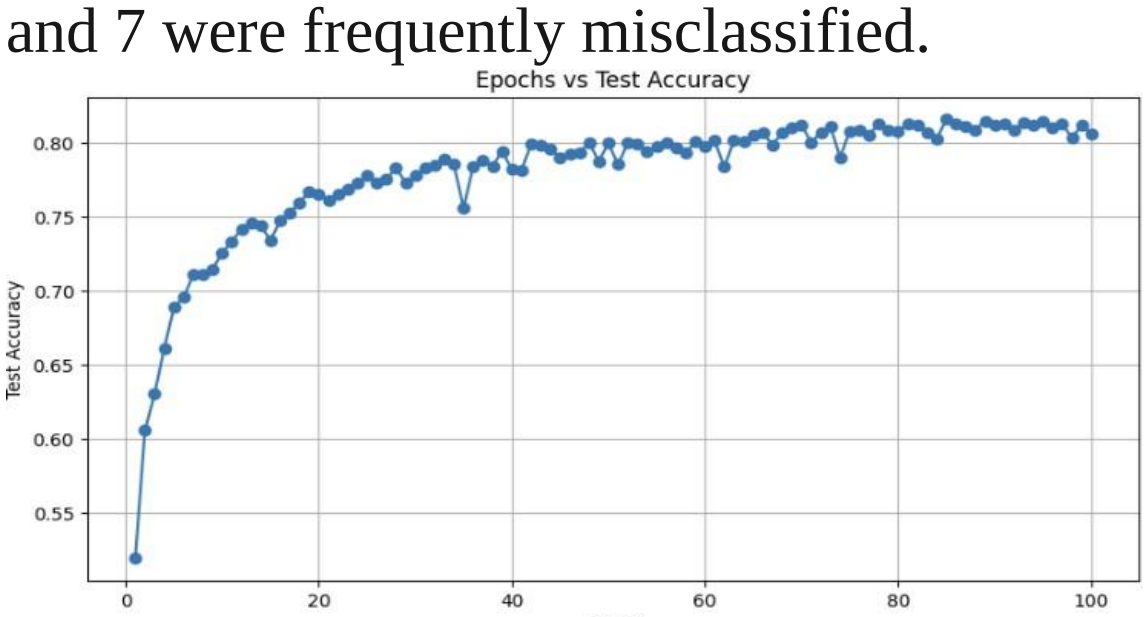

*Figure 6. Test accuracy versus training epoch for the neural network classifier in Experiment 1.*

2) k-Nearest Neighbors (KNN): The second sub- experiment evaluated a k-nearest neighbors classifier on the same dataset. Hyperparameter tuning identified an optimal value of k = 7. With this configuration, the KNN achieved an overall accuracy of 78%. This demonstrates relatively consistent performance across all labels. The corresponding confusion matrix is shown in Figure 8. Notably, the model performed better on open and closed hand gestures [12].

The overall performance may be limited by the preprocessing filter thresholds. More aggressive thresholds for the Pearson and correlation filters are expected to further reduce feature redundancy and noise. This may improve classification accuracy [7], [8].

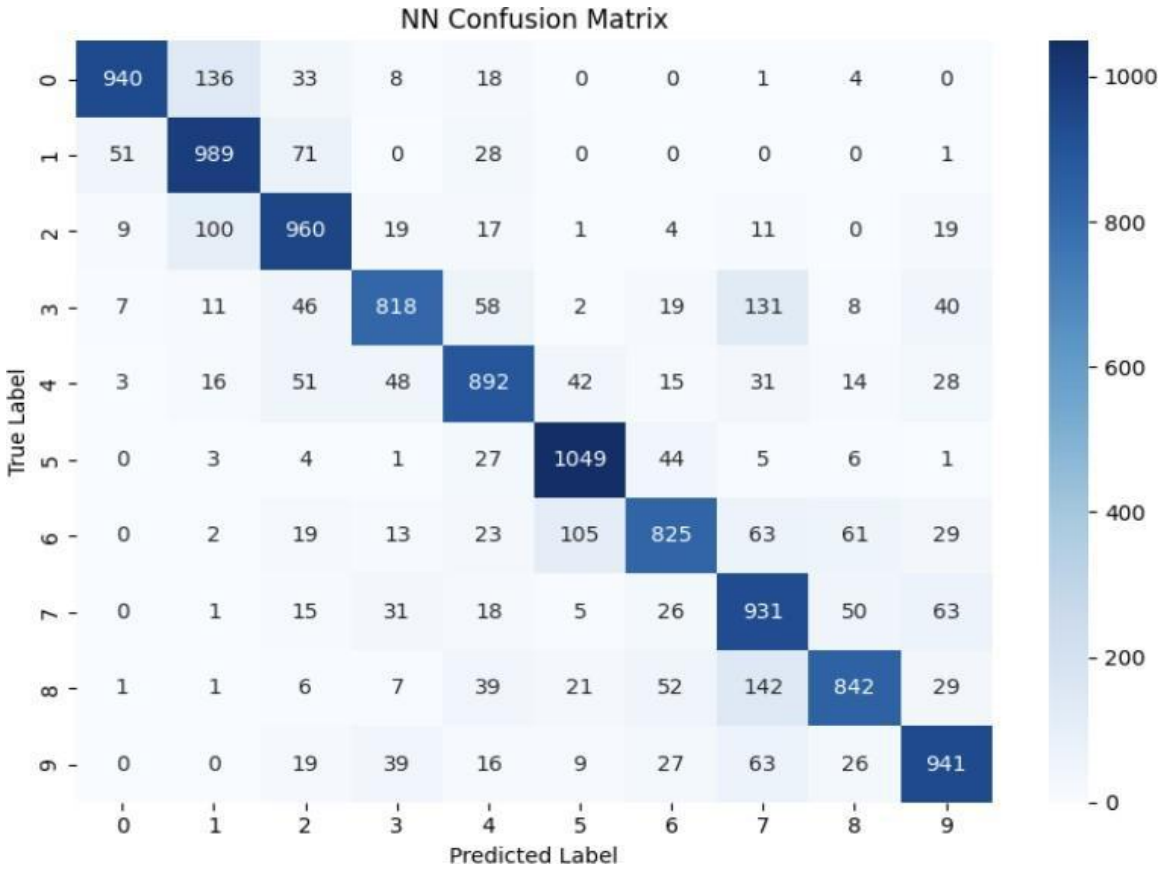

*Figure 7. Confusion matrix for the neural network classifier in Experiment 1.*

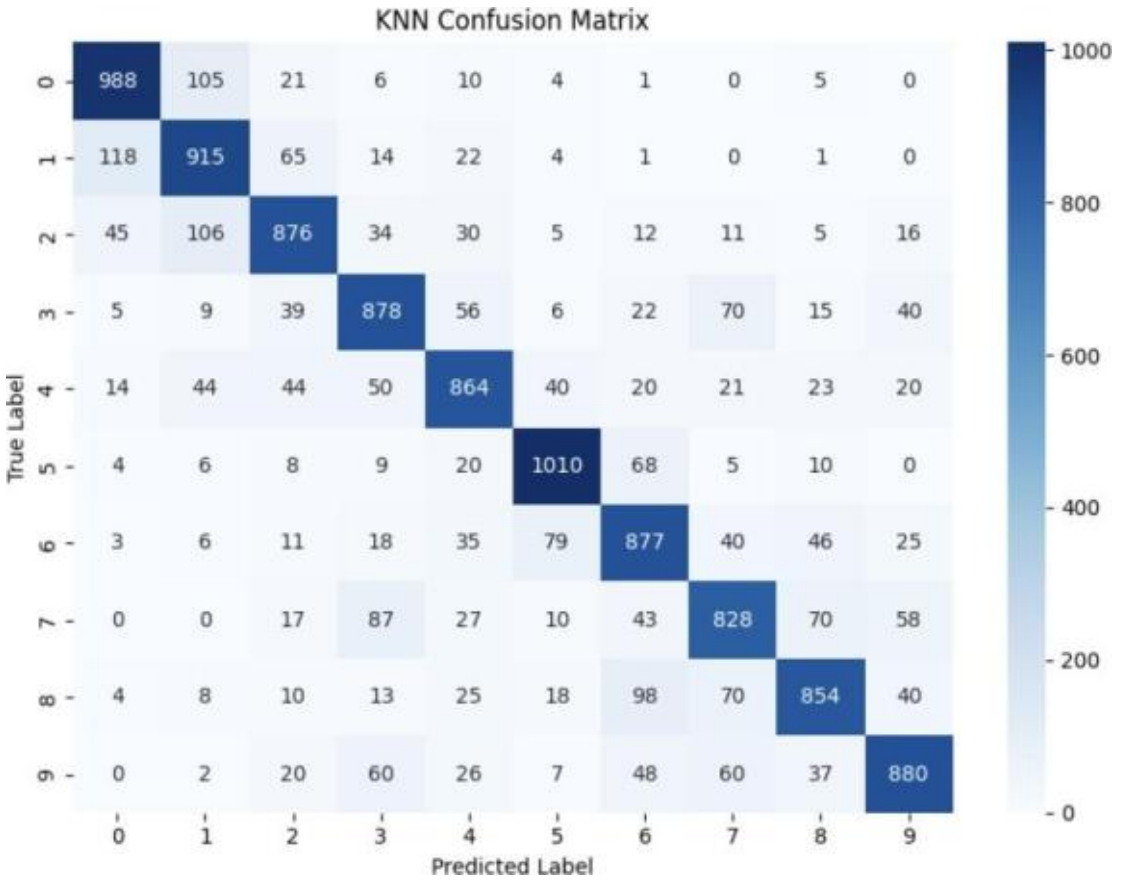

*Figure 8. Confusion matrix for the KNN classifier in Experiment 1.*

3) Support Vector Machine (SVM): The third sub experiment applied a support vector machine to the same dataset. The SVM achieved an overall accuracy of 78%, with relatively consistent performance across all gesture classes. The confusion matrix is shown in Figure 9. Similar to the KNN results, the model performed best on open and closed hand gestures. Performance is likely constrained by the current preprocessing filter thresholds. Additional optimizations of feature selection and dimensionality reduction may lead to improved results [7], [12].

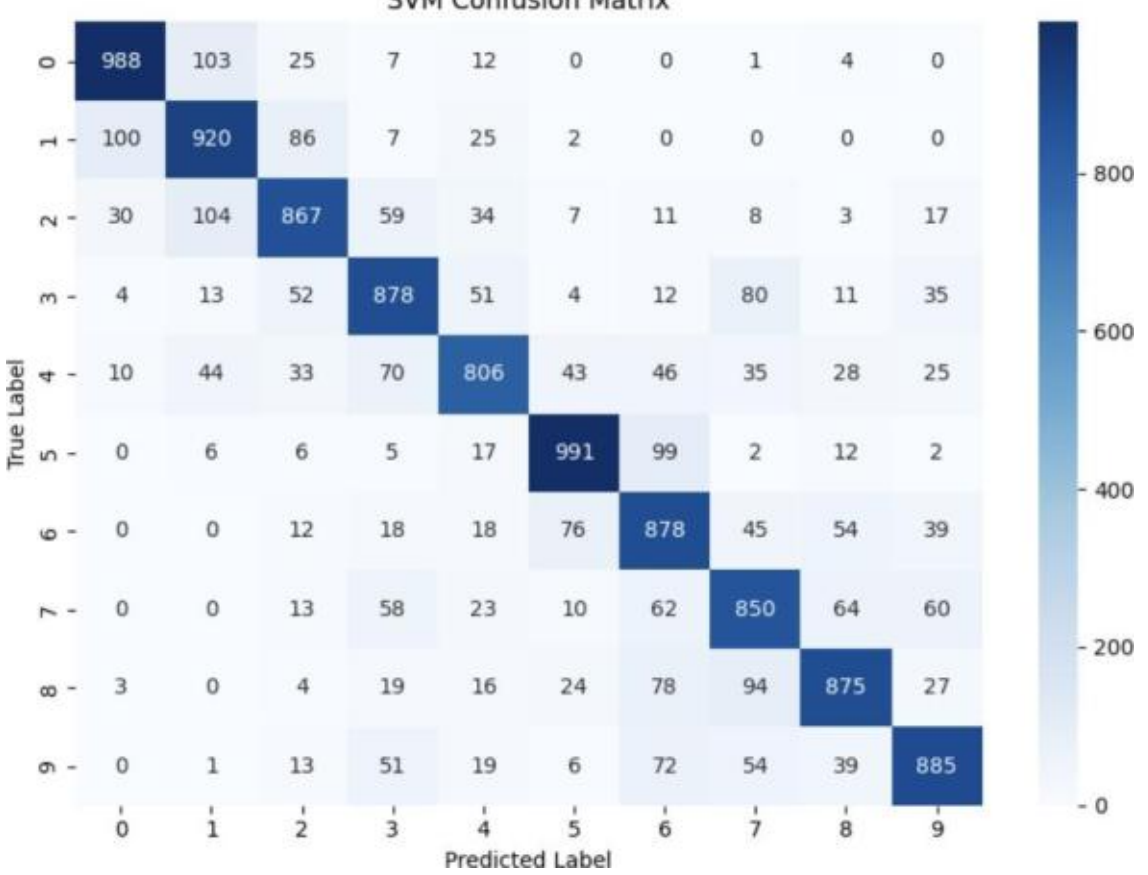

*Figure 9. Confusion matrix for the SVM classifier in Experiment 1.*

### 4.2 Experiment 2 and Results

In Experiment 2, the data was preprocessed using only the Pearson correlation filter, reducing the feature dimensionality to 26 features. The dataset was split into 80% training and 20% testing sets. Then the parameters were normalized and fit on the training set and applied to the test set to avoid data leakage [4].

1) Neural Network (NN): The first sub-experiment employed a feed-forward neural network with four hidden layers of 256, 128, 64, and 32 neurons, respectively, in addition to the input and output layers. The network used ReLU activations, a batch size of 32, the Adam optimizer with learning rate 0.001, weight decay 0.0001, and was trained for 200 epochs. This configuration achieved the highest average accuracy and F1 score of 90%, highlighting the effectiveness of Pearson filtering for NN-based models [1], [11]. Training loss plateaued around 10%, suggesting that additional temporal information from multiple channels may be needed for better generalization. Test accuracy versus epoch is shown in Figure 10, and the confusion matrix is shown in Figure 11.

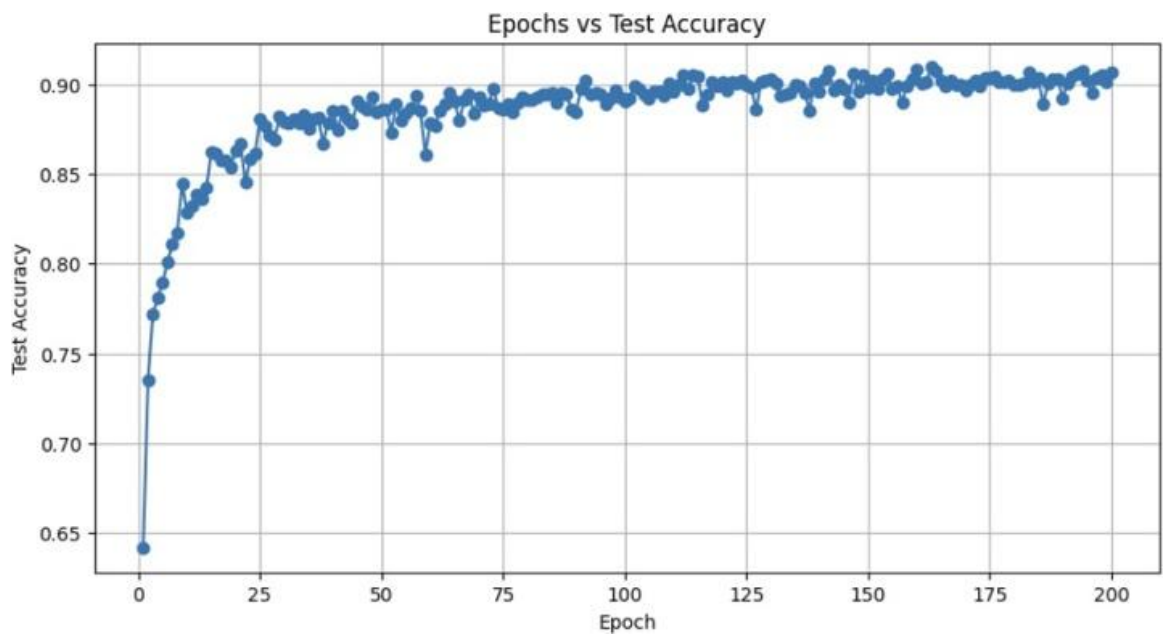

*Figure 10. Test accuracy versus training epoch for the neural network classifier in Experiment 2.*

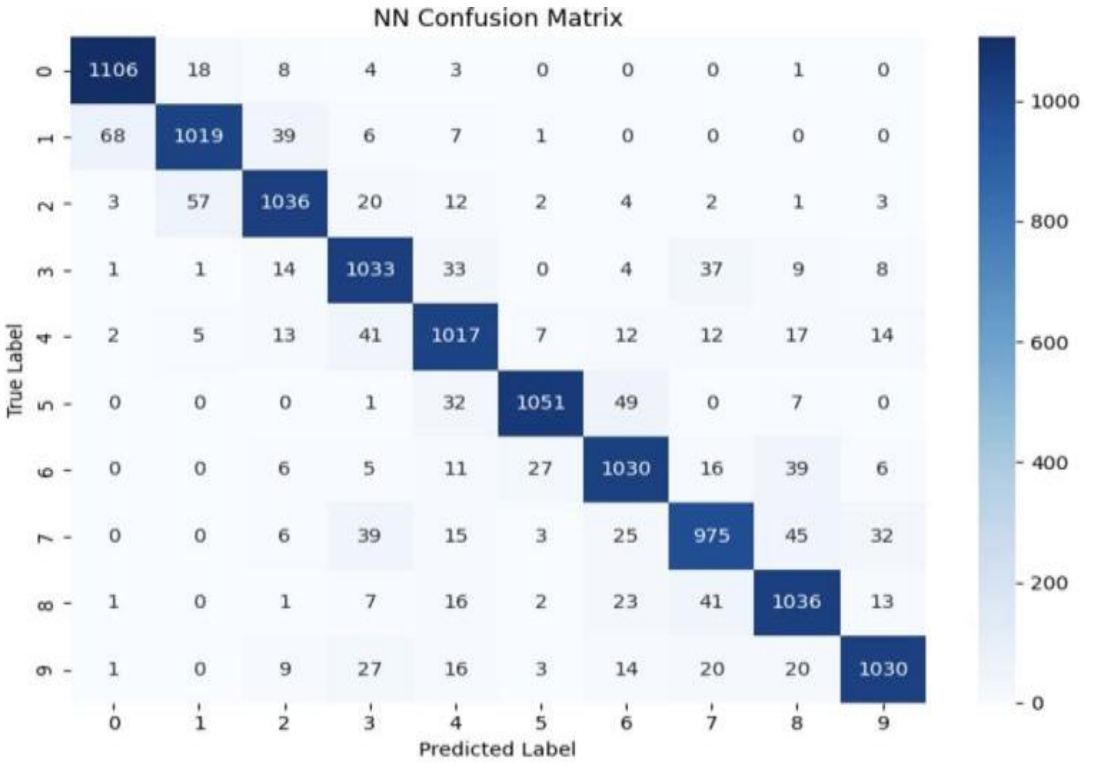

*Figure 11. Confusion matrix for the neural network classifier in Experiment 2.*

2) k-Nearest Neighbors (KNN): The second sub experiment applied a KNN classifier. Hyperparameter tuning identified an optimal k = 1. The KNN achieved an overall accuracy of 31%. This poor performance is due to the presence of highly correlated features remaining after Pearson filtering, which adversely affected distance-based classification

[7]. The confusion matrix is shown in Figure 12

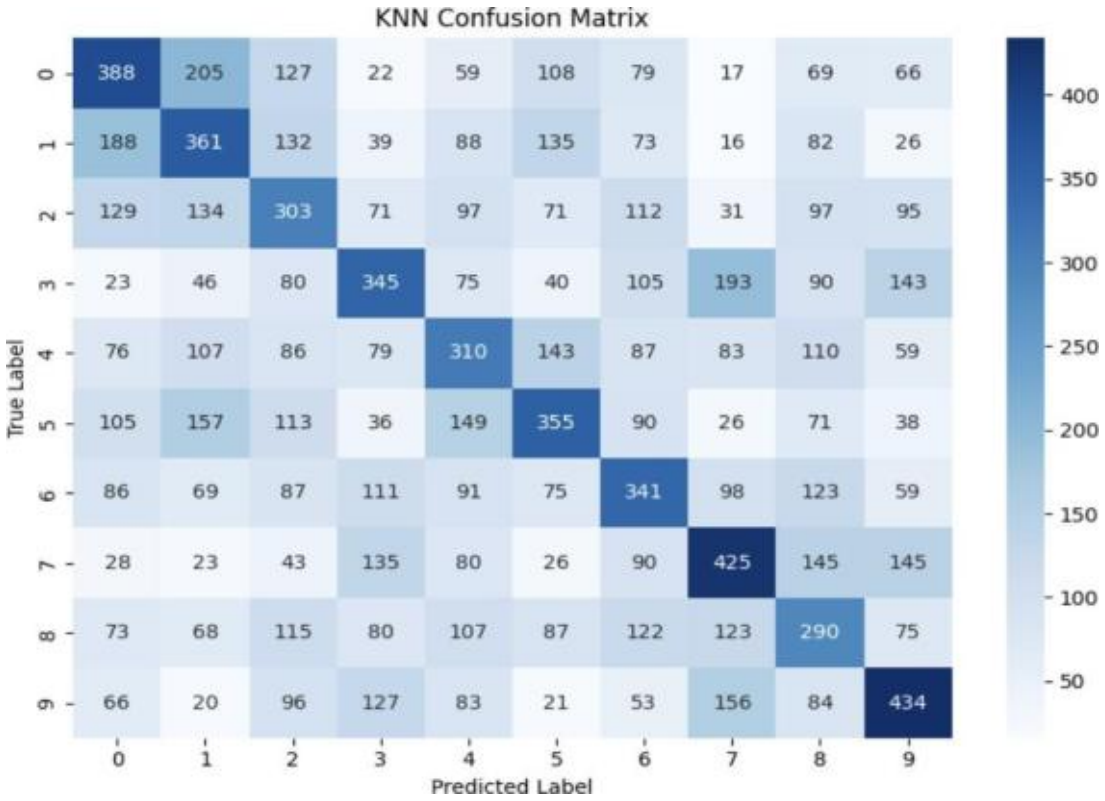

*Figure 12. Confusion matrix for the KNN classifier in Experiment 2.*

3) Support Vector Machine (SVM): The third sub- experiment applied a support vector machine to the same dataset. The SVM achieved an overall accuracy of 19%. As with the KNN, performance was constrained by the highly correlated features remaining after Pearson filtering [12]. The confusion matrix is shown in Figure 13.

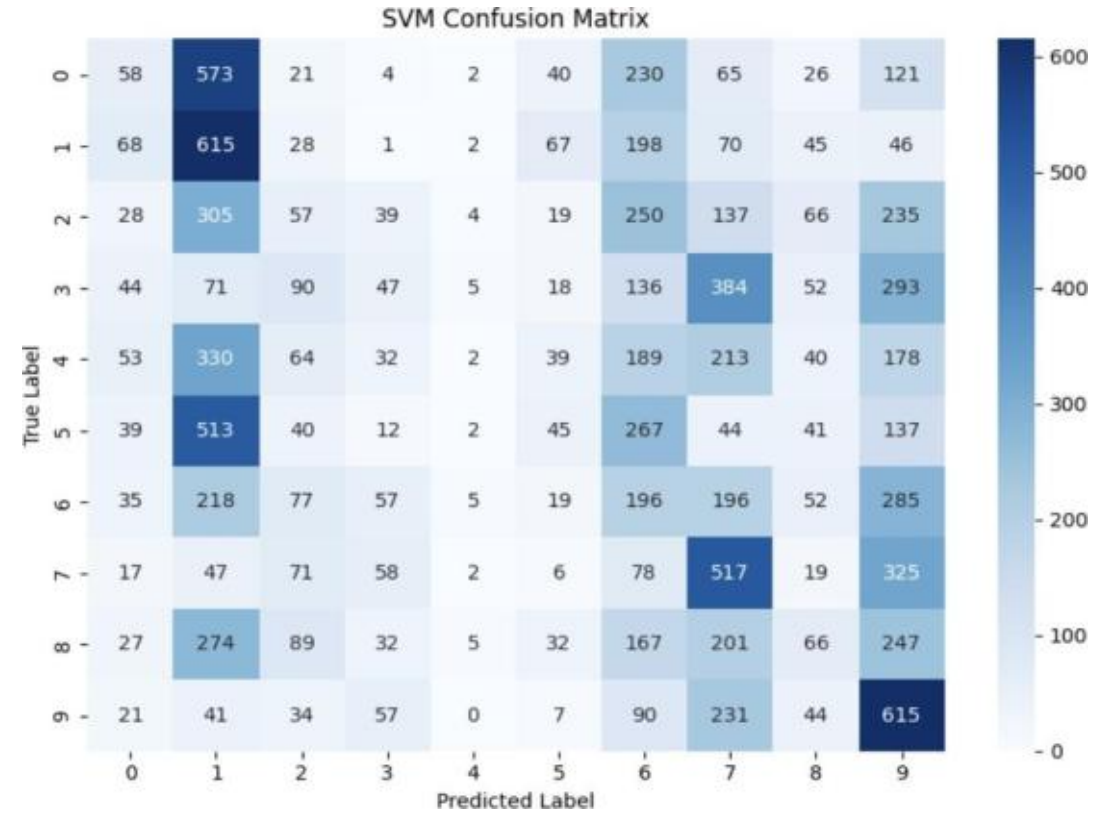

*Figure 13 Confusion matrix for the SVM classifier in Experiment 2*

## 4.3 Experiment 3 and Results

Experiment 3 applied Linear Discriminant Analysis (LDA) to further reduce the 26 Pearson-filtered features to 9 components [4]. The dataset was split 80/20 for training and testing, and normalization was applied as described previously.

1) Neural Network (NN): The NN sub-experiment employed a two hidden layer network with 64 and 32 neurons. ReLU activations, batch size of 32, Adam optimizer with learning rate 0.001, and 100 epochs were used. The network achieved an average accuracy and F1 score of 75%, plateauing at similar values, indicating the limited temporal information available from a single channel [11]. Test accuracy versus epoch is shown in Figure 14, and the confusion matrix is shown in Figure 15.

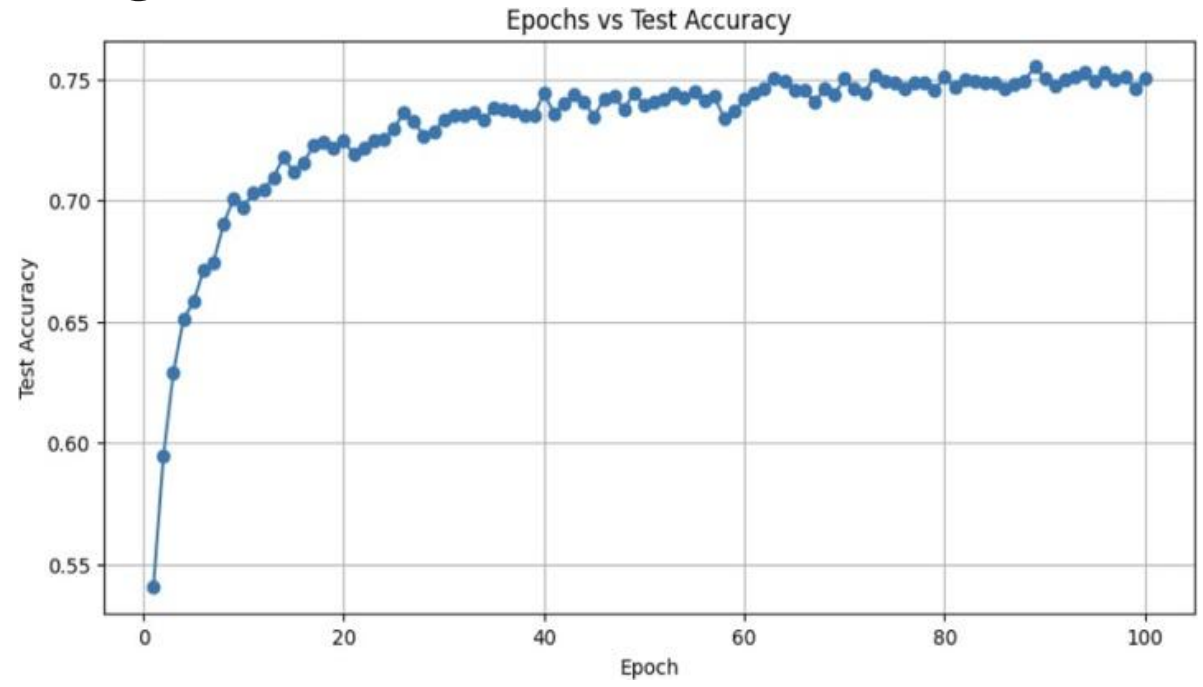

*Figure 14. Test accuracy versus training epoch for the neural network classifier in Experiment 3*

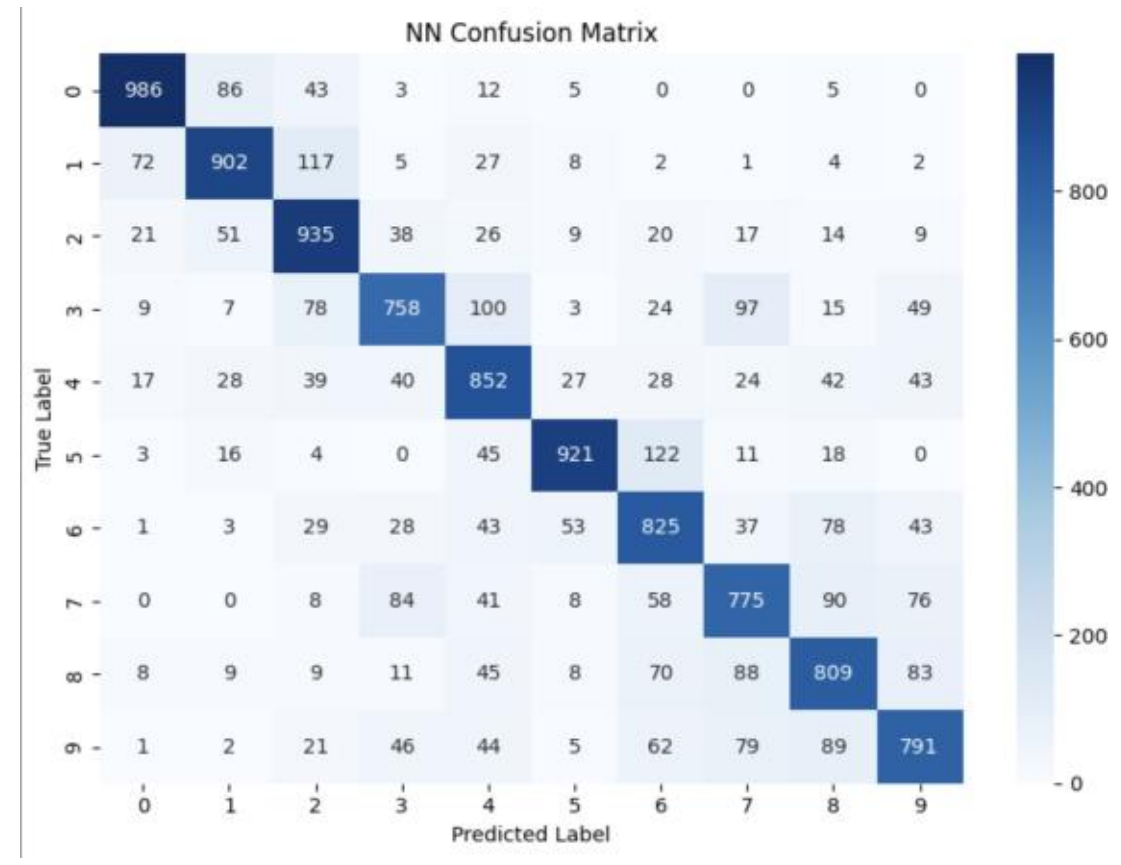

*Figure 15. Confusion matrix for the neural network classifier in Experiment 3.*

2) k-Nearest Neighbors (KNN): The KNN sub-experiment was optimized with k = 7 and

achieved an overall accuracy of 80%. The improvement compared to Experiment 2 is attributed to LDA removing redundant, highly correlated features, enhancing distance-based classification [7]. The confusion matrix is shown in Figure 16.

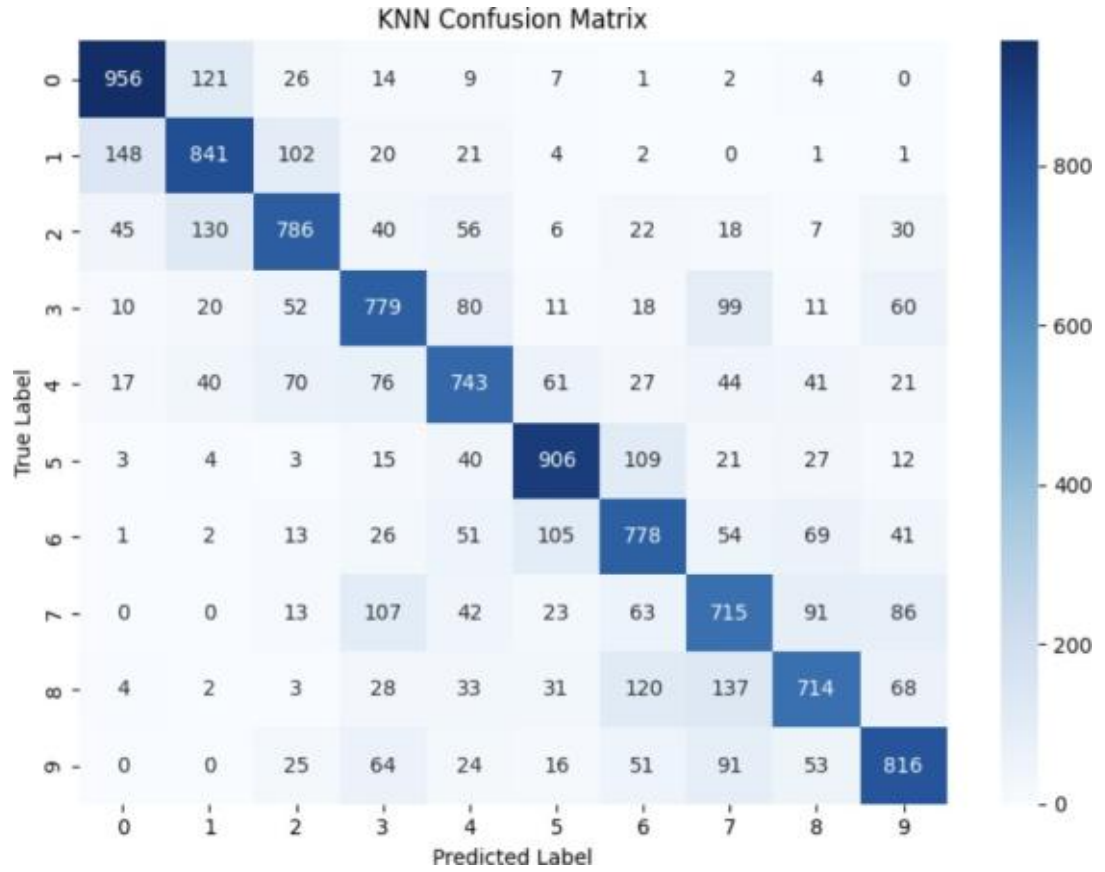

***Figure 16.*** *Confusion matrix for the KNN classifier in Experiment 3.*

3) Support Vector Machine (SVM): The SVM sub- experiment achieved an accuracy of 77%, with performance improvements similar to KNN, owing to dimensionality reduction by LDA [12]. The confusion matrix is shown in Figure 17.

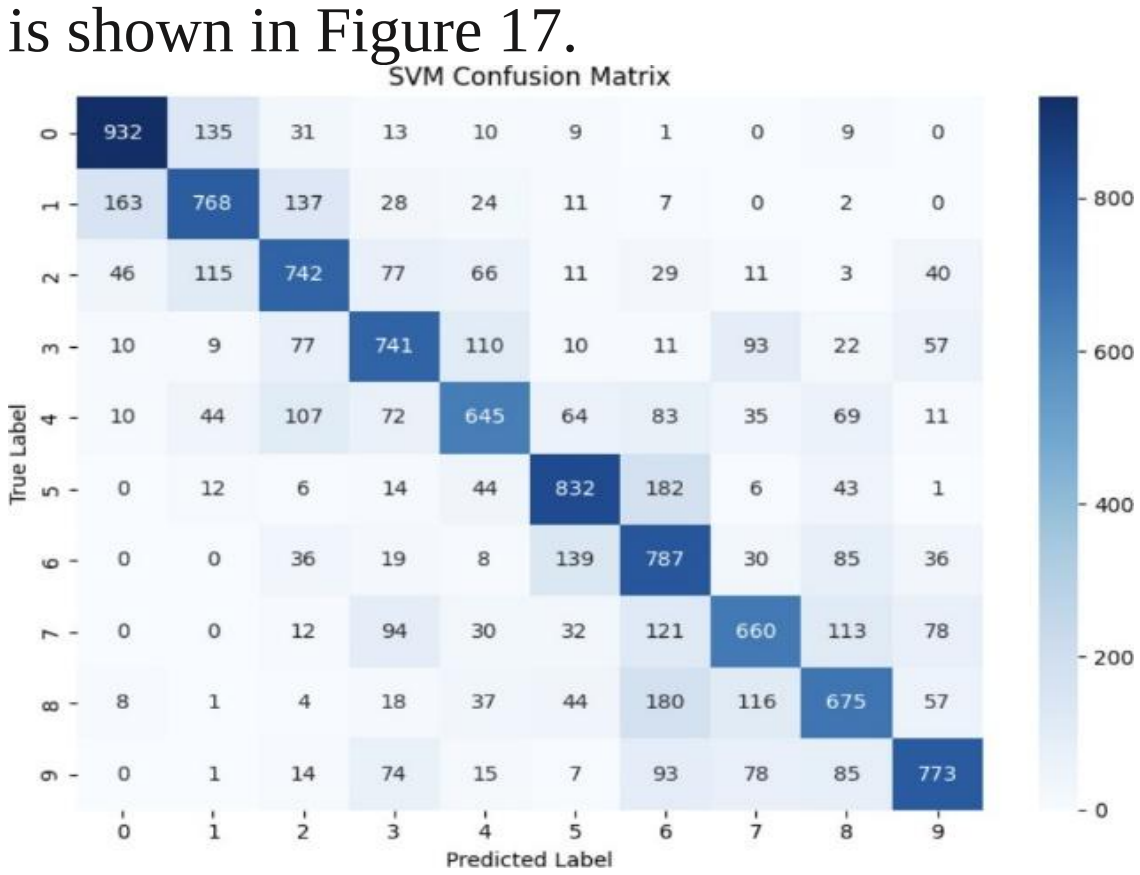

***Figure 17.*** *Confusion matrix for the SVM classifier in Experiment 3.*

## 4.4 Experiment 4 and Results

Experiment 4 applied Principal Component Analysis (PCA) to the 26 Pearson-filtered features, reducing the dimensionality to 5 components, determined using the elbow method (Figure 18) [4]. The data was split 80/20 for training and testing, and normalized accordingly.

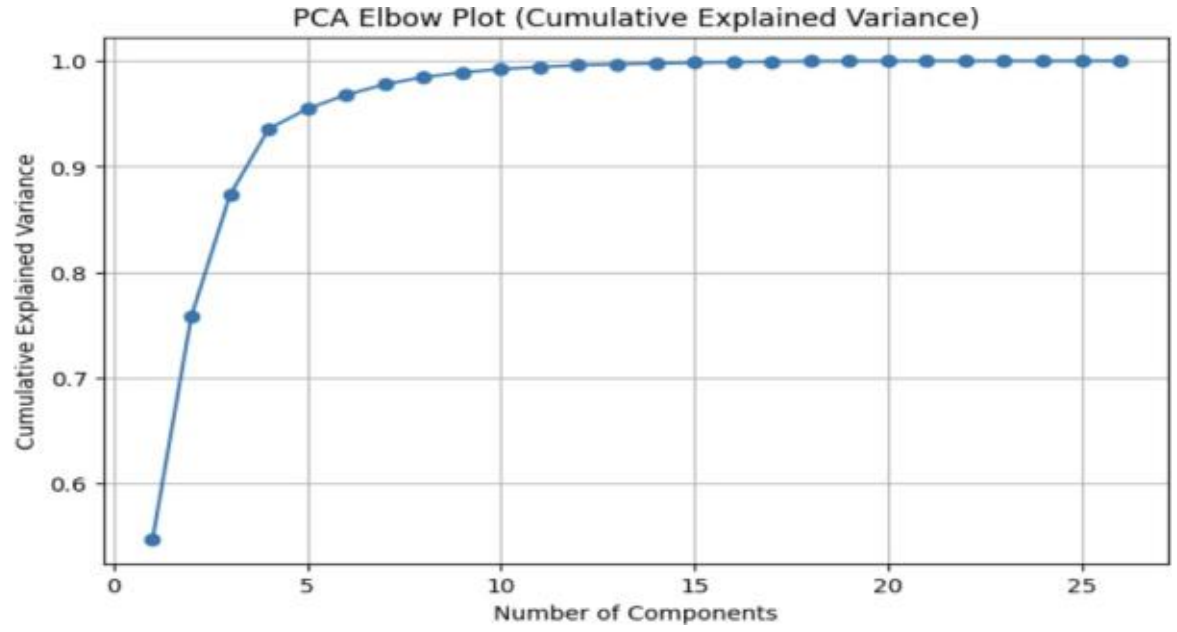

***Figure 18.*** *Elbow Test to determine optimal number of PCA components.*

`1) Neural Network (NN): The NN sub-experiment used six hidden layers with 8, 16, 32, 64, 32, and 16 neurons. The network achieved an average accuracy and F1 score of 71%, plateauing at 71%, suggesting information loss due to aggressive dimensionality reduction via PCA [11]. Test accuracy versus epoch is shown in Figure 19, and the confusion matrix is shown in Figure 20.

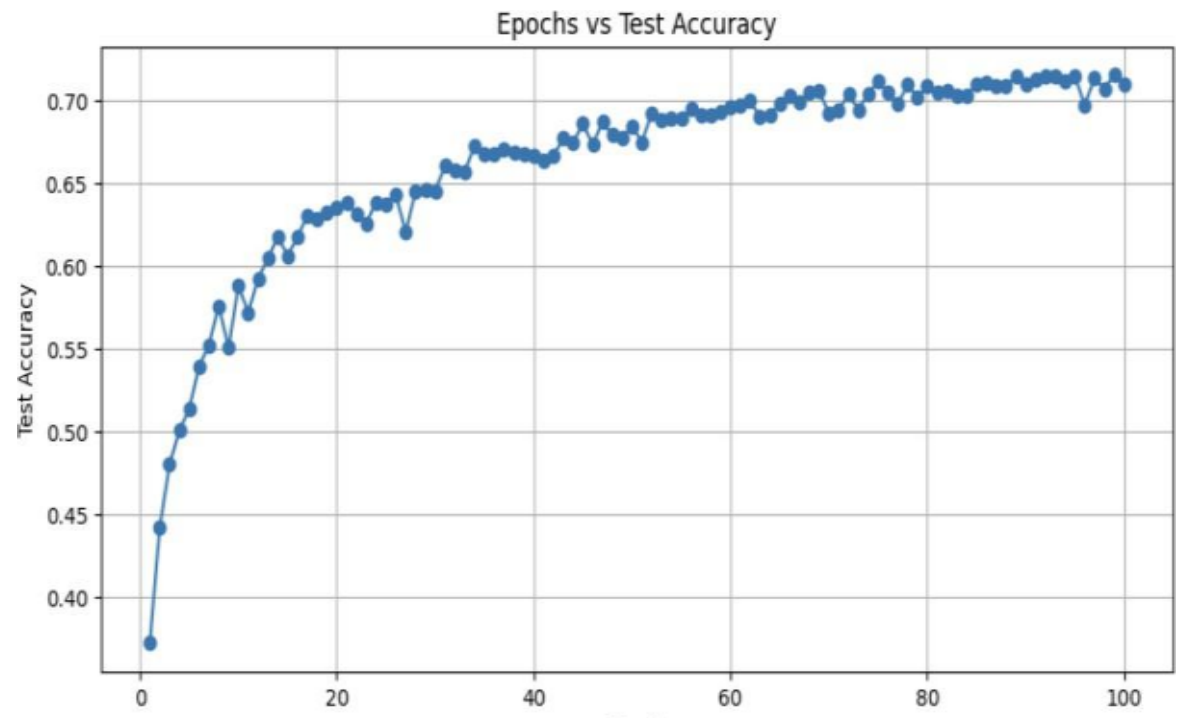

***Figure 19.*** *Test accuracy versus training epoch for the neural network classifier in Experiment 4.*

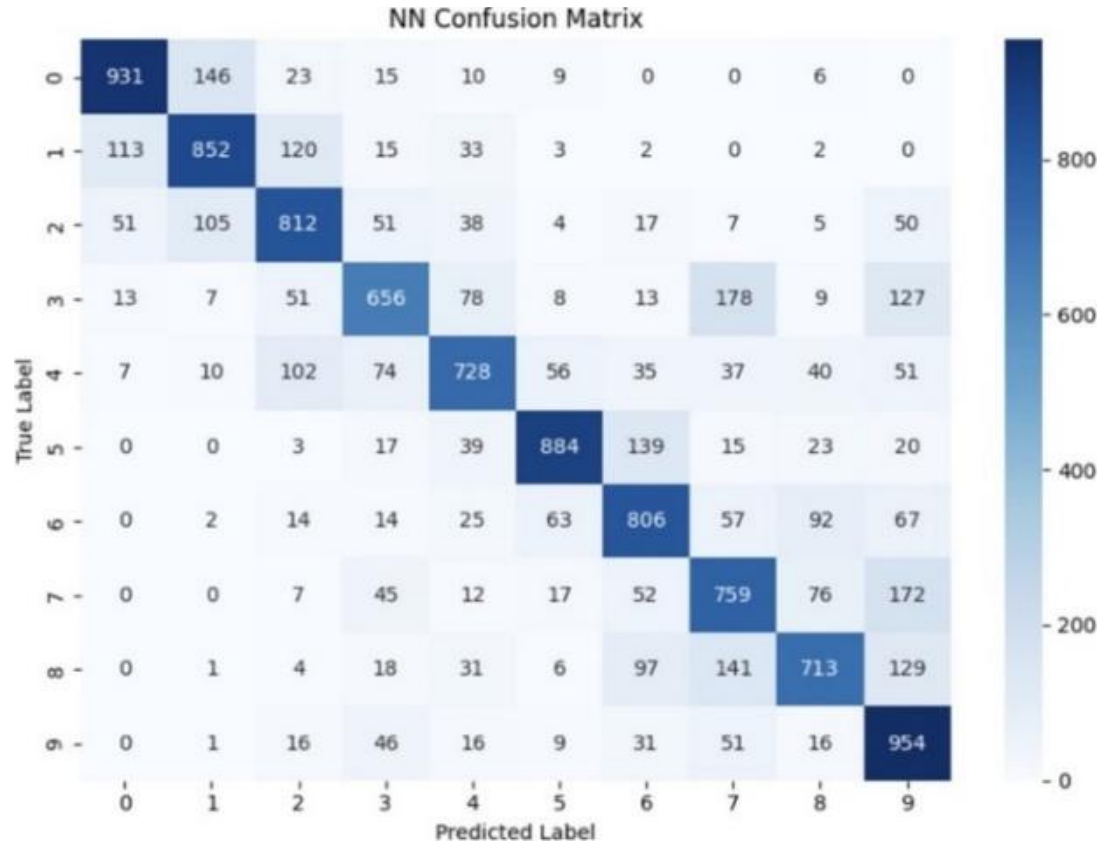

*__Figure 20.__ Confusion matrix for the neural network classifier in Experiment 4.*

2) k-Nearest Neighbors (KNN): The KNN sub- experiment, with optimized k = 9, achieved an overall accuracy of 80%. Dimensionality reduction via PCA removed highly correlated features, improving the distance-based classifier's performance [7]. The confusion matrix is shown in Figure 21.

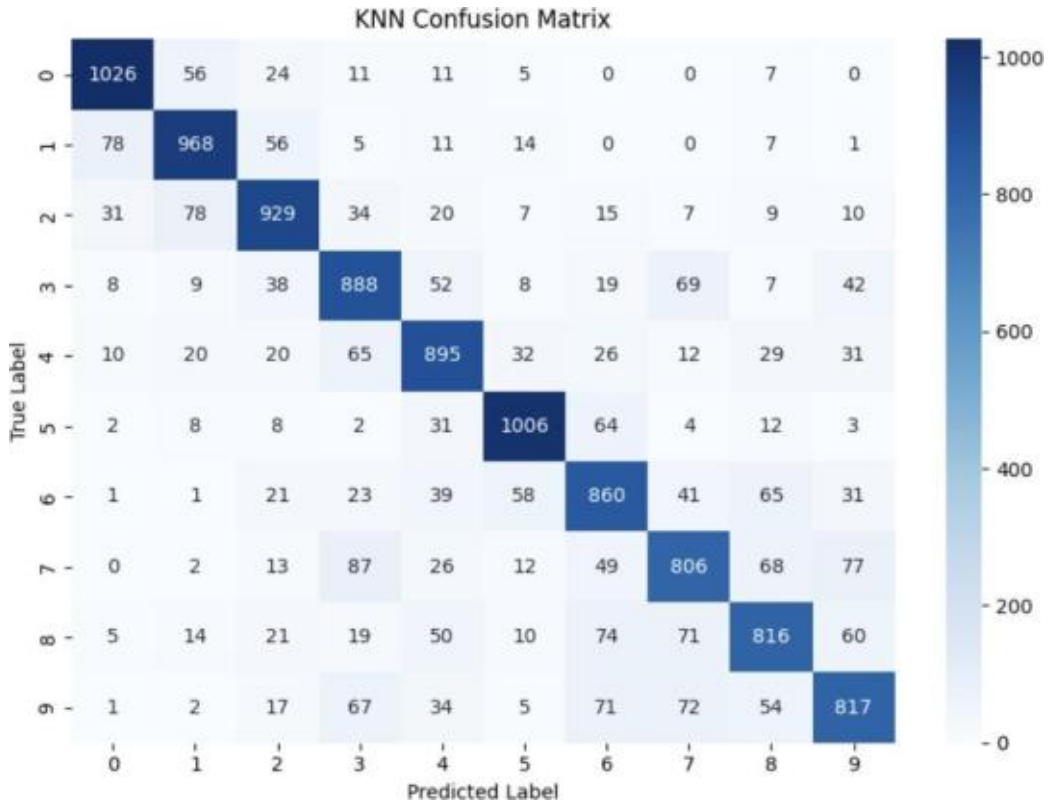

*__Figure 21.__ Confusion matrix for the KNN classifier in Experiment 4*

3) Support Vector Machine (SVM): The SVM sub- experiment achieved an ac curacy of 66%. Performance improvements compared to Experiment 2 can be attributed to PCA removing highly correlated features and reducing dimensionality [12]. The confusion matrix is shown in Figure 22.

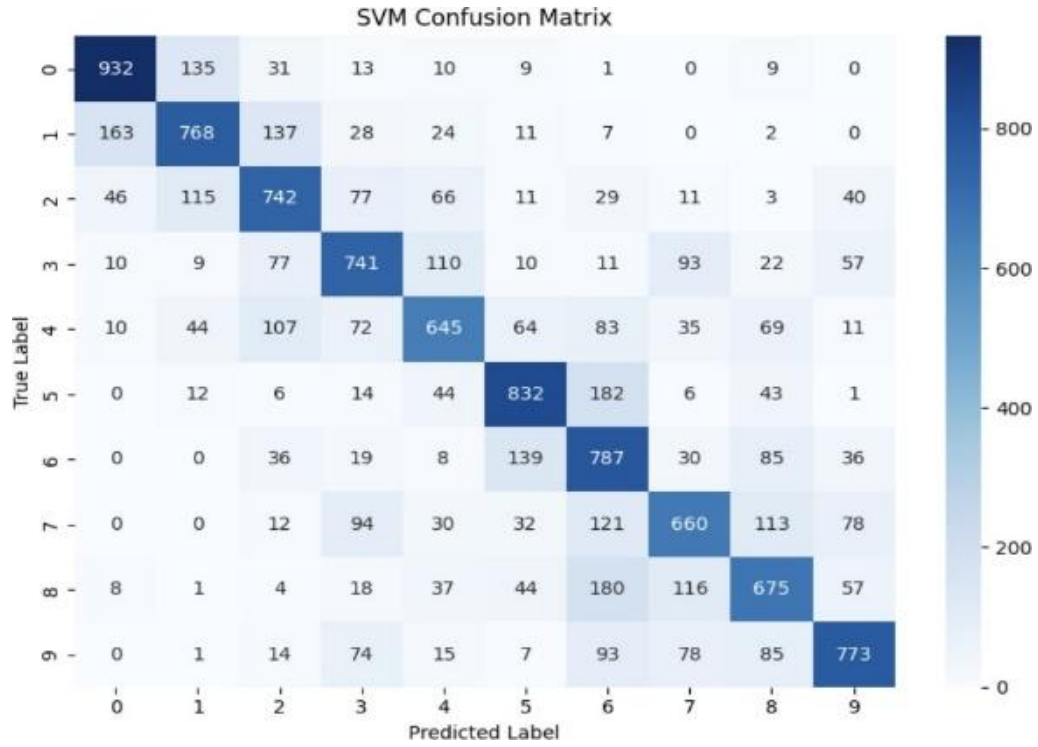

*__Figure 22.__ Confusion matrix for the SVM classifier in Experiment 4.*

# 5 Conclusion

This study demonstrates that single-channel sEMG signals, when combined with a carefully designed feature-based pipeline, can achieve competitive hand gesture classification performance. Across four experiments, we systematically evaluated the effects of correlation-based feature filtering (Pearson filter and feature correlation filter), dimensionality reduction techniques (LDA and PCA), and multiple classifier architec- tures, including neural networks (NN), k-nearest neighbors (KNN), and support vector machines (SVM).

The extracted feature set combined both time-domain and frequency-domain characteristics, enabling the models to cap- ture both amplitude-based and spectral information from the sEMG signal. The results of this study lead to the following conclusions:

Neural networks consistently outperformed KNN and SVM when trained on Pearson-filtered features, achieving up to 90% classification accuracy. This indicates that removing features with low correlation to class labels while retaining informative time- and frequency-domain features is particularly effective for NN-based models.

Distance-based classifiers, such as KNN and SVM, were more sensitive to feature redundancy. Dimensionality reduction using LDA or PCA significantly mitigated the negative impact of highly correlated features, resulting in notable performance improvements.

Aggressive dimensionality reduction (e.g., PCA to five components) led to information loss that degraded NN performance; however, the reduced feature space still benefited distance-based classifiers by minimizing redundancy and improving class separability.

Overall, these results demonstrate that a carefully designed feature-based pipeline, combining time- and frequency-domain features with Pearson correlation filtering and a compact neural network, can enable accurate single-channel sEMG hand gesture classification. This approach offers a promising pathway toward low-power, real-time, and embedded sEMG- based human–machine interface systems.